\documentclass[12pt,a4paper,onecolumn]{article}
\usepackage{natbib}
\usepackage{graphicx}
\usepackage{hyperref}
\usepackage{amsfonts} 
\usepackage{amsmath} 

\title{Interpretable non-linear dimensionality reduction using gaussian weighted linear transformation}
\author{Erik Bergh M.Sc\\  \texttt{er.bergh@gmail.com}}\date{\today}

\begin{document}

\maketitle

\begin{abstract}
\noindent
Dimensionality reduction techniques are fundamental for analyzing and visualizing high-dimensional data. With established methods like t-SNE and PCA presenting a trade-off between representational power and interpretability. This paper introduces a novel approach that bridges this gap by combining the interpretability of linear methods with the expressiveness of non-linear transformations.
The proposed algorithm constructs a non-linear mapping between high-dimensional and low-dimensional spaces through a combination of linear transformations, each weighted by Gaussian functions. This architecture enables complex non-linear transformations while preserving the interpretability advantages of linear methods, as each transformation can be analyzed independently. The resulting model provides both powerful dimensionality reduction and transparent insights into the transformed space.
Techniques for interpreting the learned transformations are presented, including methods for identifying suppressed dimensions and how space is expanded and contracted. These tools enable practitioners to understand how the algorithm preserves and modifies geometric relationships during dimensionality reduction. To ensure the practical utility of this algorithm, the creation of user-friendly
software packages is emphasized, facilitating its adoption in both academia and industry.
\end{abstract}

\section{Introduction}

Dimensionality reduction is a fundamental task in data analysis and machine learning. Its objective is to transform high-dimensional data into a more compact and meaningful representation. This process addresses critical challenges by reducing computational demands, improving visualization, and highlighting essential structures while filtering out noise. By focusing on the most relevant patterns, dimensionality reduction facilitates efficient computation and enhances the understanding of high-dimensional datasets.

Over the years, numerous techniques have been developed for dimensionality reduction, ranging from classical linear approaches to advanced non-linear methods. Principal Component Analysis (PCA) \citep{hotelling1933analysis}, one of the earliest and most influential techniques, employs a linear transformation to project data onto orthogonal axes that maximize variance. PCA is valued for its computational efficiency, scalability, and ease of interpretation, as the principal components are linear combinations of the original features. However, its reliance on linearity often limits its ability to capture non-linear relationships.

To overcome these limitations, non-linear methods have been developed to capture complex patterns in high-dimensional data. t-Distributed Stochastic Neighbor Embedding (t-SNE) \citep{van2008visualizing}, for example, is extensively used for visualizing high-dimensional data. By optimizing a model that preserves local similarities, t-SNE generates informative embeddings. Uniform Manifold Approximation and Projection (UMAP) \citep{mcinnes2018umap} leverages manifold learning to preserve both local and global data structures effectively. Other notable methods, including Isomap \citep{tenenbaum2000global} and Locally Linear Embedding (LLE) \citep{roweis2000nonlinear}, demonstrate strong representational capacity, revealing complex patterns in data. However, these methods often lack interpretability, and some, like t-SNE, cannot extend their transformations to new data without retraining. While UMAP, Isomap, and LLE can extend transformations, they come with varying degrees of computational overhead.

Deep learning have further expanded the toolkit for dimensionality reduction, with autoencoders \citep{hinton2006reducing} representing a common approach. Autoencoders, neural network architectures designed to encode data into a compressed latent space and decode it back to the original space, offer remarkable representational capacity by capturing complex relationships within data. However, they often require substantial amounts of data and lack interpretability.

Despite these advancements, a gap remains in developing methods that combine the representational power of non-linear approaches with the interpretability of linear techniques, motivating the need for new solutions.
\section{Algorithm}

\subsection{Mathematical Framework}

\subsubsection{Core Transformation}
Construct a non-linear transformation from $\mathbb{R}^{d_1}$ to $\mathbb{R}^{d_2}$ (where $d_2 < d_1$) by combining multiple linear transformations through Gaussian weighting. Each linear transformation is assigned a Gaussian function. For an input vector $\mathbf{x} \in \mathbb{R}^{d_1}$, the transformation is defined as:

\begin{equation}
    f(\mathbf{x}) = \sum_{i=1}^m w_i(\mathbf{x}) T_i(\mathbf{x})
\end{equation}
\noindent
where $m$ is the number of linear transformations with corresponding weight functions, $w_i(\mathbf{x})$ are weights, and $T_i: \mathbb{R}^{d_1} \rightarrow \mathbb{R}^{d_2}$ are linear transformations.

\subsubsection{Gaussian Weight Computation}
The weight $w_i(\mathbf{x})$ for each transformation is computed using a Gaussian function:

\begin{equation}
    g_i(\mathbf{x}) = \exp\left(-\frac{\|\mathbf{x} - \mathbf{\mu}_i\|^2}{\sigma_i^2}\right)
\end{equation}
\noindent
These weights are then normalized to sum to 1:

\begin{equation}
    w_i(\mathbf{x}) = \frac{g_i(\mathbf{x})}{\sum_{j=1}^m g_j(\mathbf{x}) + \epsilon}
\end{equation}
\noindent
where:
\begin{itemize}
 \item $\sigma_i$, the standard deviation, is optimized during training.
 \item $\epsilon$ is a small constant added for numerical stability.
 \item $\mathbf{\mu}_i \in \mathbb{R}^{d_1}$ represents the center of the $i$-th Gaussian function, initialized through random sampling from the input dataset $\mathcal{X}$. By default, these centers remain fixed during optimization.
\end{itemize}

\subsubsection{Linear Transformations}
Each $T_i$ is a linear transformation represented by a matrix $\mathbf{M}_i \in \mathbb{R}^{d_1 \times d_2}$. The transformation of a point $\mathbf{x}$ by $T_i$ is computed as:

\begin{equation}
    T_i(\mathbf{x}) = \mathbf{M}_i\mathbf{x}
\end{equation}
\noindent
\subsection{Optimization Process}

\subsubsection{Objective Function}
The algorithm minimizes the difference between pairwise distances in the original and transformed spaces. For a dataset $\mathcal{X} = \{\mathbf{x}_1, ..., \mathbf{x}_n\}$, the loss function is:

\begin{equation}
    \mathcal{L} = \frac{1}{N} \sum_{i,j} \left(\|\mathbf{x}_i - \mathbf{x}_j\| - \|f(\mathbf{x}_i) - f(\mathbf{x}_j)\|\right)^2
\end{equation}
\noindent
where $N$ is the number of considered pairs.

\subsubsection{Training Procedure}
The optimization process consists of two phases:

\paragraph{Initialization Phase}
\begin{itemize}
    \item The Gaussian centers $\mathbf{\mu}_i$ are initialized through random sampling from the input dataset.
    \item The standard deviations, $\sigma_i$, are initialized to unity.
    \item Transformation matrices $\mathbf{M}_i$ are initialized using a random distribution.
    \item Compute pairwise distances between all points in the original space.
\end{itemize}

\paragraph{Optimization Phase}
For each iteration:
\begin{enumerate}
    \item Forward pass: compute transformed points $f(\mathbf{x}_i)$ for all data points.
    \item Compute pairwise distances in the transformed space.
    \item Update parameters using gradient descent, such as the Adam optimizer \citep{kingma2014adam}.
    \begin{itemize}
        \item Standard deviations $\sigma_i$.
        \item Transformation matrices $\mathbf{M}_i$.
        \item Optionally, Gaussian centers $\mathbf{\mu}_i$ if enabled.
    \end{itemize}
\end{enumerate}

\subsubsection{Distance Computation Optimization}
To improve computational efficiency, the algorithm considers only the $k$ nearest neighbors during loss function computation:

\begin{equation}
    \mathcal{L}_k = \frac{1}{N} \sum_{i} \sum_{j \in \mathcal{N}_k(i)} \left(\|\mathbf{x}_i - \mathbf{x}_j\| - \|f(\mathbf{x}_i) - f(\mathbf{x}_j)\|\right)^2
\end{equation}
\noindent
where $\mathcal{N}_k(i)$ represents the $k$ nearest neighbors of point $\mathbf{x}_i$ in the original space. 

The optimization process stops based on the chosen termination criterion, such as exceeding a predefined patience threshold or reaching the maximum number of epochs.

\section{Interpretability}
\begin{figure}[h]
    \centering
    \includegraphics[width=\linewidth]{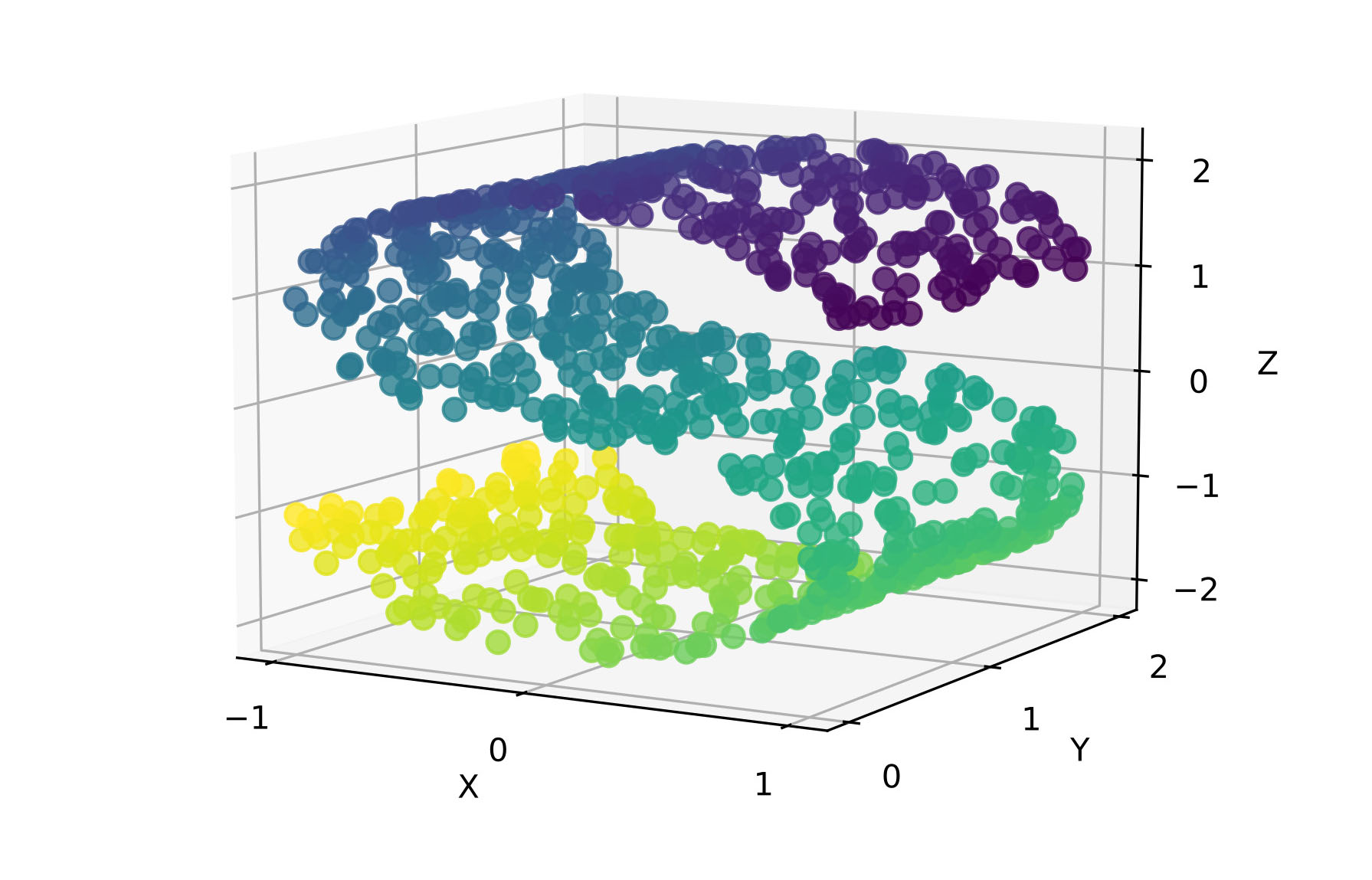}
    \caption{Visualization of a dataset generated using \textit{Scikit-learn's make\_s\_curve}.}
    \label{fig:s}
\end{figure}

To showcase the interpretability of a trained model, the algorithm is applied to a 3-dimensional S-shaped dataset created using \textit{Scikit-learn's make\_s\_curve} function. A total of 1,000 data points are generated, as illustrated in Figure \ref{fig:s}. For details, refer to \url{https://scikit-learn.org/stable/modules/generated/sklearn.datasets.make_s_curve.html}. This section does not present an exhaustive list of possible interpretability techniques for the proposed algorithm. Instead, it presents those techniques implemented at \url{https://github.com/erikbergh/interpretable_dim_reduction}, as of the time of writing.

\begin{figure}[h]
    \centering
    \includegraphics[width=\linewidth]{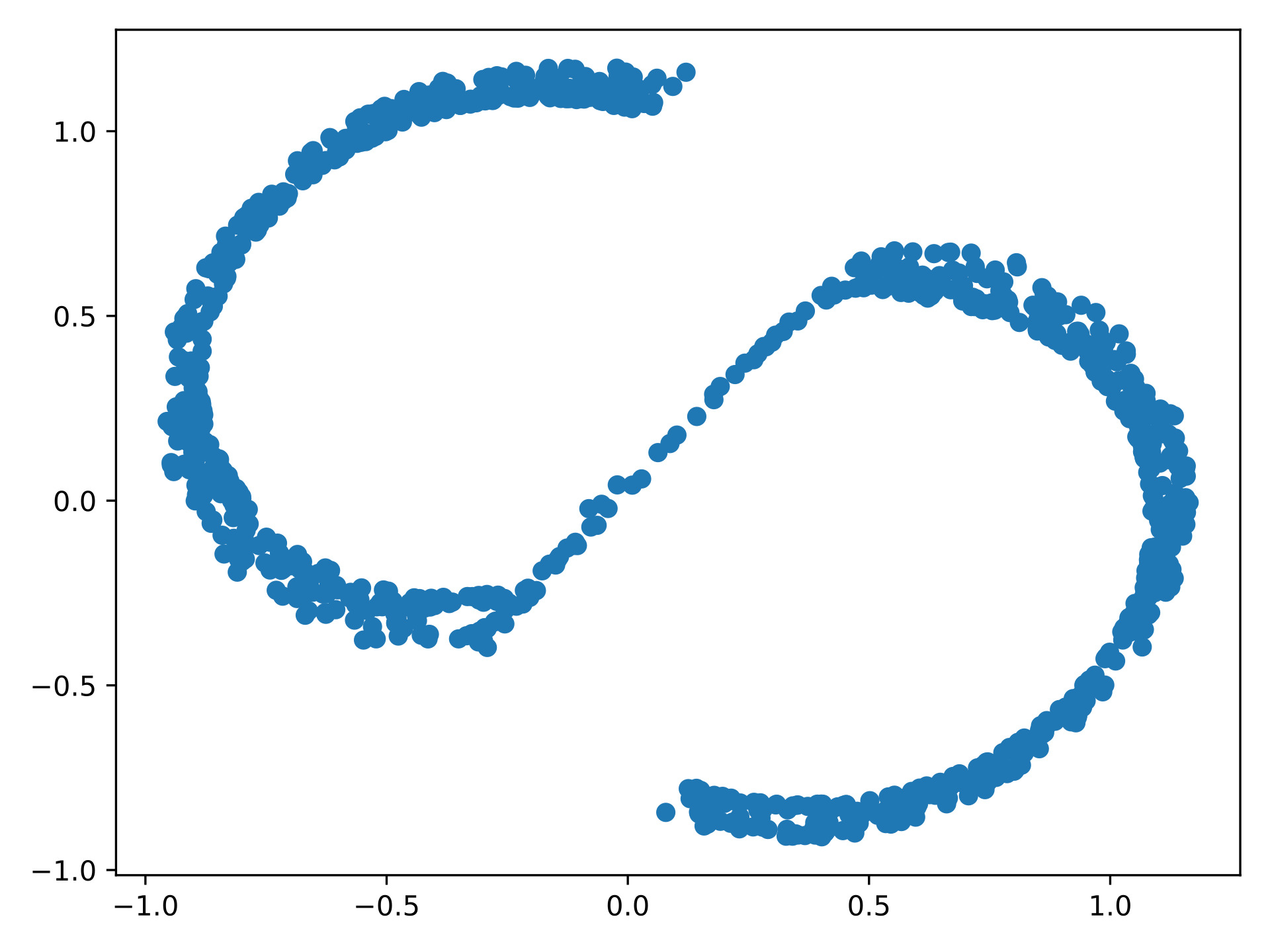}
    \caption{The data points visualized in the reduced 2D space.}
    \label{fig:reduced}
\end{figure}

The algorithm employs 100 Gaussian functions and accounts for all pairwise distances. The output dimensionality is reduced to two, and the training is performed over 2,000 epochs. The resulting dimensionality reduction is depicted in Figure \ref{fig:reduced}.

The reconstruction error, which measures how well the distances are preserved, is given by:  
\begin{equation}
\text{error} = \frac{\sum_{i,j} \left| ||\mathbf{x}_i-\mathbf{x}_j||_2 - ||f(\mathbf{x}_i)-f(\mathbf{x}_j)||_2 \right|}{\sum_{i,j} ||\mathbf{x}_i- \mathbf{x}_j||_2}
\end{equation}  
\noindent
The reduction shown in Figure \ref{fig:reduced} has a reconstruction error of 0.45 (rounded to two decimal places), where a value of zero indicates a perfect preservation of distances.

\subsection{Influence by dimension}  
By comparing Figure \ref{fig:s} and Figure \ref{fig:reduced}, it is evident that the y-axis contributes less prominently in the reduced dimensions. Such comparisons are challenging for most datasets. The influence of each dimension can be calculated as follows:  

\begin{equation}
\bar{w} = \frac{1}{N} \sum_{i} \left( \frac{\sum_{k} |M_{ijk}|}{\sum_{j,k} |M_{ijk}|} \right)
\end{equation}  
\noindent

Here, \(i\) denotes the index of the linear transformations, \(j\) represents the column index, and \(k\) corresponds to the row index. For this dimensionality reduction, \(\bar{w} = [0.40, 0.25, 0.35]\) (rounded to two decimal places), confirming that the y-dimension is less represented in the reduction.  

\begin{figure}[h]
    \centering
    \includegraphics[width=\linewidth]{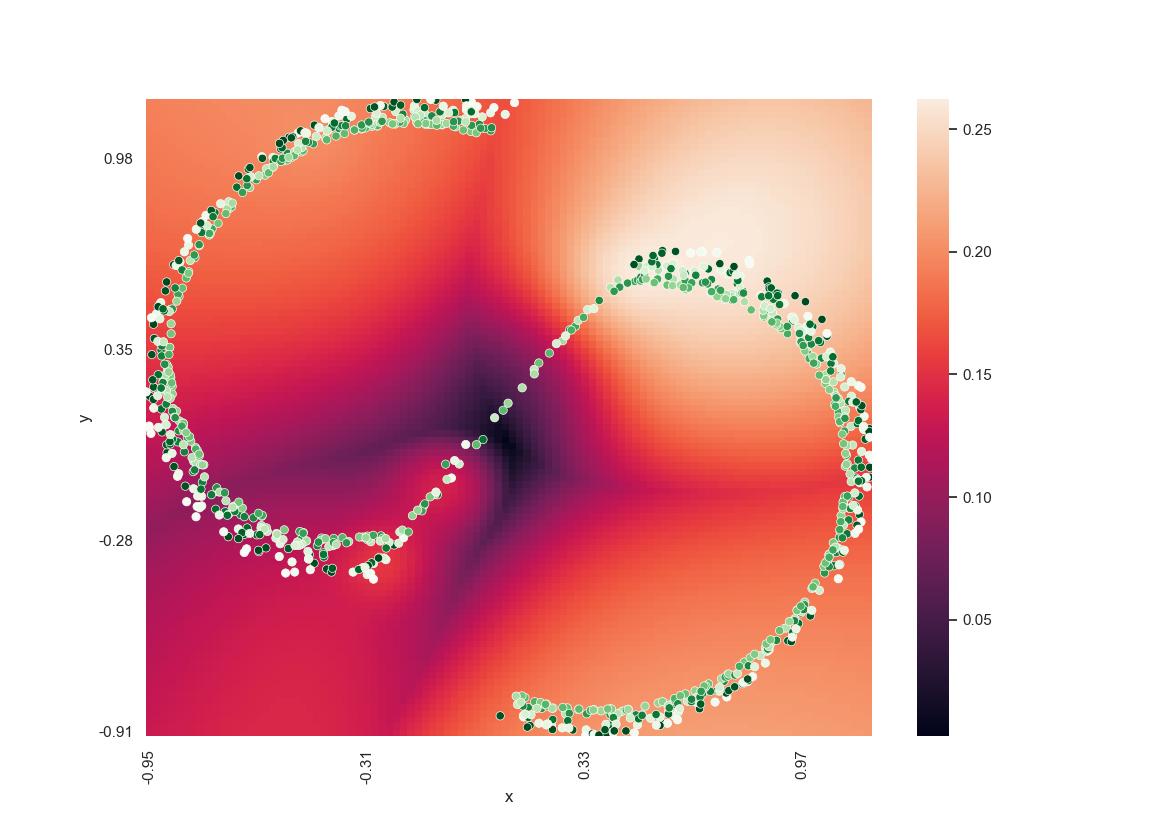}
    \caption{The values \(\bar{w}(p)_{j=2}\), corresponding to the original y-dimension, are shown as the background. The data points in the reduced space are plotted on top, with the point colors corresponding to their original y-values.}
    \label{fig:influence}
\end{figure}

Due to the non-linear nature of the dimensionality reduction, these values vary across the space. Understanding how the influence of each dimension changes spatially can provide additional insights. By calculating:  

\begin{equation}
\bar{w}(p) = \frac{1}{N} \sum_{i} \left( \frac{\sum_{k} |w(p)_i M_{ijk}|}{\sum_{j,k} |w(p)_i M_{ijk}|} \right)\label{eq:wbar}
\end{equation}  

where \(p\) represents a point in the reduced-dimensional space, \(\bar{w}(p)\) is evaluated for points forming a mesh grid over the reduced space. The values of \(\bar{w}(p)_{j=2}\), corresponding to the original y-dimension, are shown as the background in Figure \ref{fig:influence}, with the data points in the reduced space plotted on top. The color of the points corresponds to their original y-values.

Figure \ref{fig:influence} illustrates that the influence of the original y-axis is minimal at the center of the S-shape. In contrast, other regions show a greater influence from the original y-axis, although this influence remains less pronounced compared to the contributions of the x and z axes.

\subsection{Influence Skewness}
It is of interest to know if the reduced space is skewed to represent or under-represent one or more original dimensions. By calculating the variance of $\bar{w}(p)$ in Equation \eqref{eq:wbar} for each point \(p\), we obtain the variance of influence across the different dimensions. 

\begin{figure}[h]
    \centering
    \includegraphics[width=\linewidth]{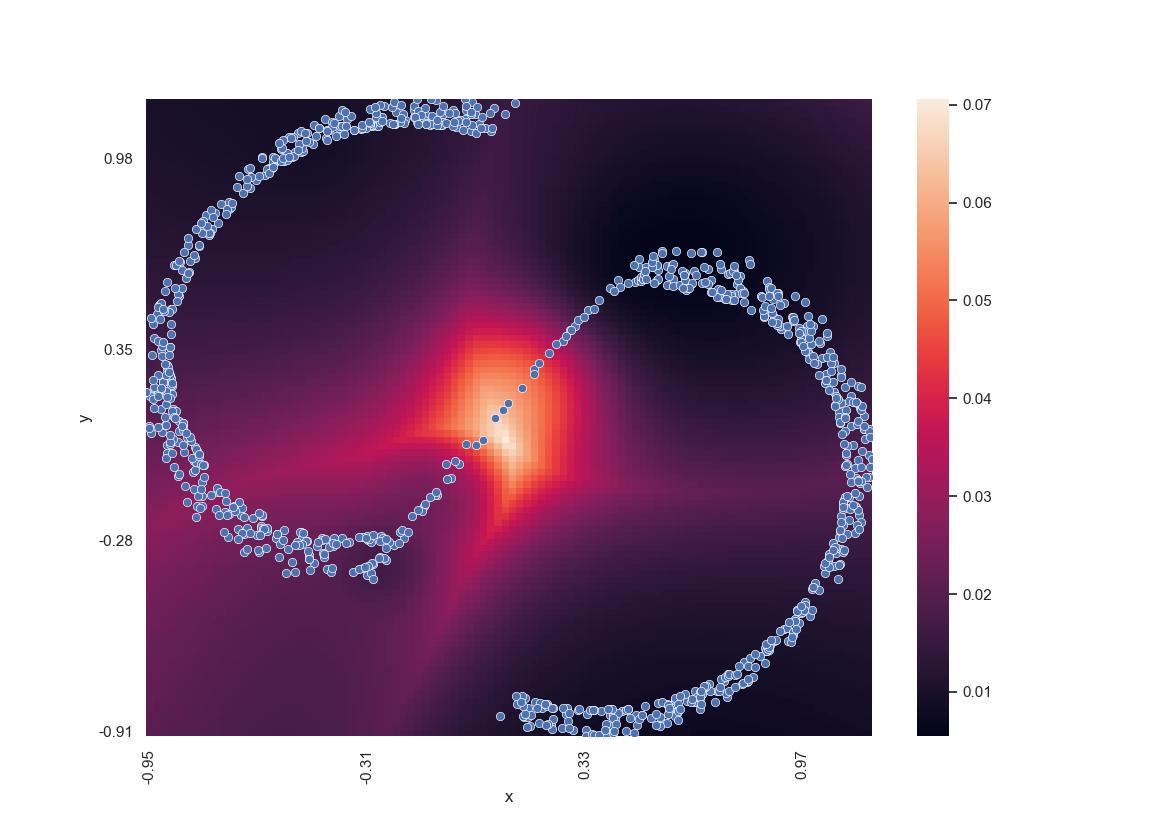}
    \caption{The variance $Var(\bar{w}(p))$ over the mesh grid is plotted as the background. Data points in the reduced space are plotted on top.}
    \label{fig:skew}
\end{figure}

In Figure \ref{fig:skew}, this variance is plotted as the background, with the data points in the reduced space shown on top.

As shown in Figure \ref{fig:skew}, the maximum skewness occurs at the center of the shape. This is directly related to the minimal influence of the original dimension \(y\), as illustrated in Figure \ref{fig:influence}.

\subsection{Expansion and Contraction of Space}
It is useful to determine whether the reduced space is expanded or contracted at a given point. This analysis helps compare the relative distances between neighboring points, such as a nearby pair versus a distant pair. 

The expansion or contraction at a point \(p\) is quantified using: 
\begin{equation}
N(p) = \left\| \sum_{i} w(p)_i M_i \right\|_2,
\end{equation}
\noindent
where \(N(p)\) represents the norm of the transformation at point \(p\). A value of \(N(p) > 1\) indicates that the space is expanded at \(p\), while \(N(p) < 1\) indicates contraction. 

\begin{figure}[h]
    \centering
    \includegraphics[width=\linewidth]{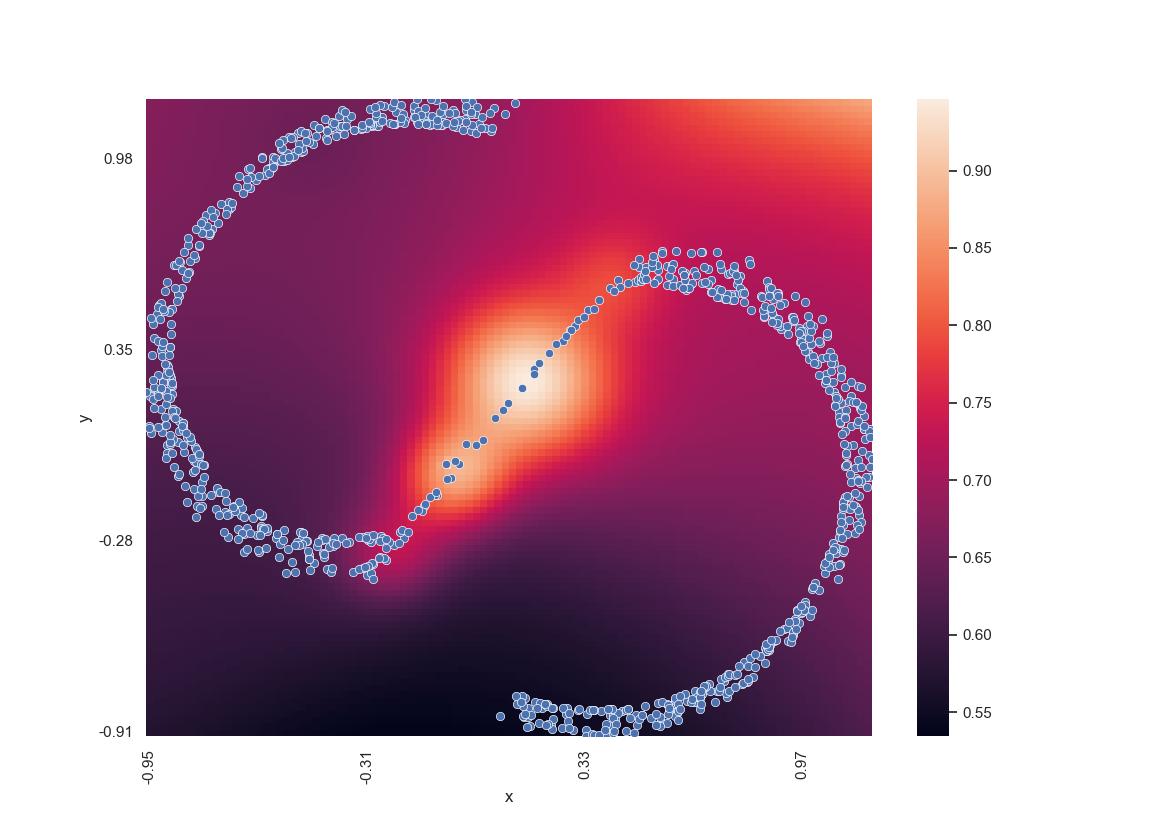}
    \caption{The values of \(N(p)\) calculated over the mesh grid are shown as the background. Data points in the reduced space are plotted on top.}
    \label{fig:norm}
\end{figure}

To visualize this, \(N(p)\) is computed over a set of points forming a mesh grid in the reduced space. In Figure \ref{fig:norm}, \(N(p)\) is shown as the background, with data points in the reduced space plotted on top.

As shown in Figure \ref{fig:norm}, the background indicates that the space is contracted everywhere. The contraction is minimal at the center of the S-shape.

\section{Discussion and Conclusion}
The proposed algorithm demonstrates both representational power and interpretability, addressing key limitations of traditional dimensionality reduction techniques. Unlike t-SNE, which is constrained to the dataset on which it is trained, the proposed algorithm can extend its transformations to new data points with no additional computational cost compared to processing a data point from the training set. Furthermore, the algorithm's representational capacity enables it to capture complex patterns and relationships within the data, offering richer representations than linear methods like PCA.

However, the algorithm presents several challenges. Like t-SNE, it requires substantial computational resources for training, limiting its scalability for very large datasets or resource-constrained environments. Additionally, the risk of converging to local minima can affect result quality and stability. The algorithm’s robustness remains an open question, particularly regarding whether the risk of suboptimal representation increases with dataset complexity.

Future work should examine its computational complexity on large and complex datasets and compare performance with methods such as t-SNE. Another key area is benchmarking its representational power against other algorithms. Additionally, improving model interpretability is essential. While the algorithm shows promise in this regard, systematic methods for extracting insights and articulating relationships in a user-friendly manner are needed. This is especially crucial for high-dimensional datasets where in-depth investigation of each dimension is unfeasible. Developing intuitive yet effective interpretation techniques without sacrificing insights would significantly enhance the algorithm’s usability, making it a valuable tool for analyzing complex data.

In conclusion, this algorithm combines representational power with interpretability, bridging the gap between popular techniques like t-SNE and PCA. By creating intuitive and meaningful software packages for interpretation, the algorithm has the potential to become a widely used tool in both academic and industrial settings.

\section{Resources}
A Python implementation of the proposed method is available at \url{https://github.com/erikbergh/interpretable_dim_reduction}. The repository includes the algorithm, a minimal working example, and interpretability demonstrations.

\section{Conflict of interest}
There are no conflicts of interest known to the author.

\bibliography{references}

\end{document}